%% file: emnlp-ijcnlp-2019.tex
\documentclass[11pt,a4paper]{article}
\usepackage[hyperref]{emnlp-ijcnlp-2019}
\usepackage{times}
\usepackage{latexsym}

\usepackage{url}
\usepackage{xcolor}
\usepackage{url}            %
\usepackage{amsfonts}       %
\usepackage{nicefrac}       %
\usepackage{microtype}      %
\usepackage{subcaption}
\usepackage{multirow}
\usepackage{amsmath}
\usepackage{amsthm}
\usepackage{mathtools}
\usepackage{verbatim}
\usepackage{amssymb}
\usepackage{colortbl}
\usepackage{xspace}
\usepackage{tikz}
\usepackage{afterpage}
\usepackage{dblfloatfix}
\usepackage{enumitem}
\usepackage{placeins}
\usepackage{booktabs}
\usepackage{url}

\aclfinalcopy %

\newcommand{\T}{^\top}
\newcommand{\y}{\ensuremath{y_\textrm{true}}\xspace}
\newcommand{\onehot}[1]{\ensuremath{\mathbf{e}_{#1}}\xspace}
\DeclareMathOperator*{\maximize}{maximize}

\newcommand{\lbound}[2]{\ensuremath{\underline{#1}_{#2}}\xspace}
\newcommand{\ubound}[2]{\ensuremath{\overline{#1}_{#2}}\xspace}
\newcommand{\lowerbound}[2]{\ensuremath{\lbound{#1}{#2}(\delta)}\xspace}
\newcommand{\upperbound}[2]{\ensuremath{\ubound{#1}{#2}(\delta)}\xspace}

\newcommand{\z}{\ensuremath{\mathbf{z}\xspace}}
\newcommand{\x}{\ensuremath{\mathbf{x}\xspace}}
\newcommand{\cvec}{\ensuremath{\mathbf{c}\xspace}}
\newcommand{\p}{\ensuremath{\mathbf{p}\xspace}}
\newcommand{\ignore}[1]{}

\newcommand{\pswrite}[1]{#1}
\newcommand{\jwwrite}[1]{#1}
\newcommand{\svcomment}[1]{#1}

\title{Achieving Verified Robustness to Symbol Substitutions\\ via Interval Bound Propagation}

\author{
\renewcommand*{\thefootnote}{\fnsymbol{footnote}}
Po-Sen Huang$^\dag$ \quad
 Robert Stanforth$^{\dag}\footnotemark[4]$ \quad
Johannes Welbl$^{\ddagger}\footnotemark[4]$ \quad
Chris Dyer$^\dag$\\
\textbf{Dani Yogatama}$^{\dag}$ ~
\textbf{Sven Gowal}$^{\dag}$ ~ 
\textbf{Krishnamurthy Dvijotham}$^{\dag}$ ~ \textbf{Pushmeet Kohli}$^{\dag}$\\ 
\\
$^\dag$DeepMind ~
$^{\ddagger}$University College London  \\
 {\small \tt  \{posenhuang, stanforth, cdyer, dyogatama, sgowal, dvij, pushmeet\}@google.com} ~ \\\tt {\small \{j.welbl\}@cs.ucl.ac.uk~}
 }
\date{}

\date{}

\begin{document}
\maketitle
\renewcommand*{\thefootnote}{\fnsymbol{footnote}}
\footnotetext[3]{Work done during an internship at DeepMind.}
\footnotetext[4]{Equal contribution.}
\renewcommand*{\thefootnote}{\arabic{footnote}}

\begin{abstract}
  Neural networks are part of many contemporary NLP systems, yet their empirical successes come at the price of vulnerability to adversarial attacks.
  Previous work has used adversarial training and data augmentation to partially mitigate such brittleness, but these are unlikely to find worst-case adversaries due to the complexity of the search space arising from discrete text perturbations.
  In this work, we approach the problem from the opposite direction: to formally verify a system's robustness against a predefined class of adversarial attacks.
  We study text classification under synonym replacements or character flip perturbations. %
  We propose modeling these input perturbations as a simplex and then using Interval Bound Propagation -- a formal model verification method.\footnote{The source code is available at \url{https://github.com/deepmind/interval-bound-propagation/tree/master/examples/language/}}
  We modify the conventional log-likelihood training objective to train models that can be efficiently verified, which would otherwise come with exponential search complexity.
  The resulting models show only little difference in terms of nominal accuracy, but have much improved verified accuracy under perturbations and come with an efficiently computable formal guarantee on worst case adversaries.
\end{abstract}

\begin{figure}[t]
\centering
\includegraphics[width=.99\linewidth]{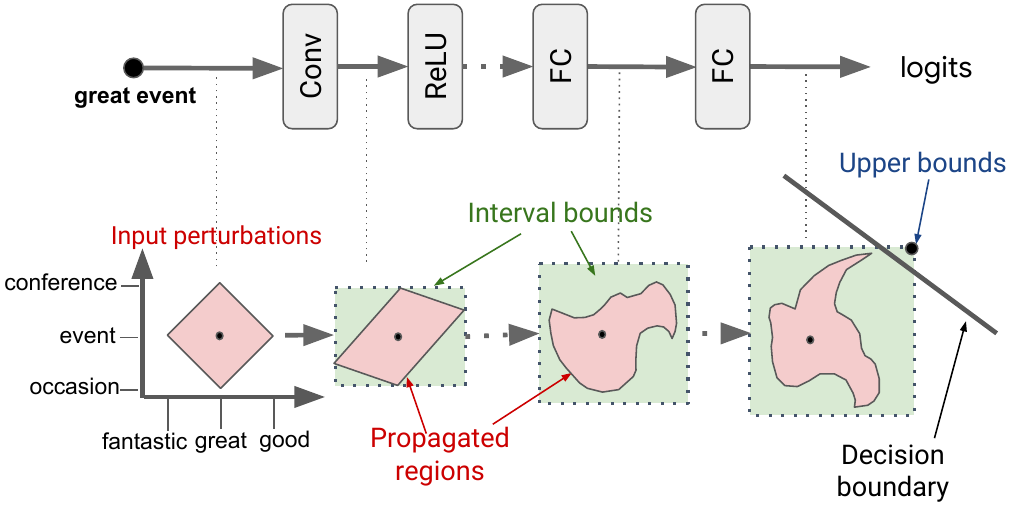}
\caption{{
Illustration of verification \pswrite{with the input simplex and } Interval Bound Propagation.
From the left, input perturbations define the extreme points of a simplex (in red, projected to 2D here) around the statement ``great event'' that is propagated through a model. 
At each layer, this shape deforms itself, but can be bounded by axis-parallel bounding boxes, which are propagated similarly.
Finally, in logit space, we can compute an upper bound on the worst-case specification violation (e.g., prediction changes). %
}}
\label{fig:diagram}
\end{figure}

\section{Introduction}

Deep models have been shown to be vulnerable against adversarial input perturbations \cite{szegedy2013intriguing,kurakin2016adversarial}. 
Small, semantically invariant input alterations can lead to drastic changes in predictions, leading to poor performance on adversarially chosen samples.
Recent work \cite{jia2017adversarial,Belinkov_Bisk2017, Ettinger17} also exposed the vulnerabilities of neural NLP models, e.g.~with small character perturbations \cite{ebrahimi2018hotflip} or paraphrases \cite{ribeiro2018semantically,iyyer2018syntactically}.
These adversarial attacks highlight often unintuitive model failure modes and present a challenge to deploying NLP models.

Common attempts to mitigate the issue are adversarial training \cite{ebrahimi2018hotflip} and data augmentation \cite{Belinkov_Bisk2017, Yitong-2017}, which lead to improved accuracy on adversarial examples. 
However, this might cause a false sense of security, as there is generally no guarantee that stronger adversaries could not circumvent defenses to find other successful attacks \cite{carlini2017adversarial,athalye2018obfuscated,uesato2018adversarial}.
Rather than continuing the race with adversaries, \emph{formal verification} \cite{baier2008principles, barrett2018satisfiability,katz2017reluplex} offers a different approach: 
it aims at providing provable guarantees to a given model specification. 
In the case of adversarial robustness, such a specification can be formulated as prediction consistency under \emph{any} altered -- but semantically invariant -- input change.

In this paper, we study verified robustness\pswrite{, i.e., providing a certificate that for a given network and test input, no attack or perturbation under the specification can change predictions,} using the example of text classification tasks, Stanford Sentiment Treebank (SST) \cite{socher2013recursive} and AG News \cite{zhang2015character}.
The \pswrite{\emph{specification} against which we verify is that a text classification model should preserve its prediction under character (or synonym) substitutions in a character (or word) based model.}
We \pswrite{propose modeling these input perturbations as a simplex and then using} Interval Bound Propagation (IBP) \citep{IBP, mirman2018differentiable, dvijotham2018training} to compute worst case bounds on specification satisfaction, as illustrated in Figure~\ref{fig:diagram}.
Since these bounds can be computed efficiently, we can furthermore derive an auxiliary objective for models to \emph{become} verifiable.
The resulting classifiers are efficiently verifiable and improve robustness
on adversarial examples, while maintaining comparable performance in 
terms of nominal test accuracy.

The contributions of this paper are twofold: %
\begin{itemize}
    \item \svcomment{To the best of our knowledge, this paper is the first to introduce} verification and verifiable training for neural networks in natural language processing  (\S{\ref{sec:methods}}). 
    \item Through a series of experiments (\S{\ref{sec:experiments}}), we demonstrate
    (a) the effectiveness of modeling input perturbations as a simplex and using simplex bounds with IBP for training and testing,
    (b) the weakness of adversarial training under exhaustive verification,
    (c) the effects of perturbation space on the performance of different methods, and
    (d) the impact of using GloVe and counter-fitted embeddings on the IBP verification bounds. 
\end{itemize}

\section{Related Work}
\label{sec:relatedwork}
\paragraph{Adversarial Examples in NLP.}
Creating adversarial examples for NLP systems requires identifying semantically invariant text transformations to define an input perturbation space.
In this paper, \pswrite{given our specification,} we study word- and character-level \emph{HotFlip} attacks \cite{ebrahimi2018hotflip} -- which consist of character and synonym replacements -- on text classification tasks. We compare our verifiable approach to other defenses including adversarial training \cite{goodfellow2014explaining} and data augmentation \cite{Yitong-2017, Belinkov_Bisk2017}. 
\pswrite{Note that some existing adversarial perturbations such as syntactically controlled paraphrasing \cite{iyyer2018syntactically}, exploiting backtranslation systems \cite{ribeiro2018semantically}, output-lengthening attacks \cite{Wang_2019_CVPR}, or using targeted keyword attack \cite{seq2sick} are beyond the specification in this paper. 
}

\paragraph{Formal Verification of Neural Networks.}
Formal verification provides a provable guarantee that models are consistent with a \emph{specification} for all possible model inputs.
Previous work can be categorised into complete methods that use Mixed-Integer Programming (MIP)  \citep{bunel2017piecewise,cheng2017maximum} or Satisfiability Modulo Theory (SMT) \citep{katz2017reluplex,carlini2017ground}, and  incomplete methods that solve a convex relaxation of the verification problem \citep{weng2018towards,wong2018provable,wang2018formal}.
Complete methods perform exhaustive enumeration to find the worst case. 
Hence, complete methods are expensive and difficult to scale, though they
provide exact robustness bounds.
Incomplete methods provide loose robustness bounds, but can be more scalable and used inside the training loop for training models to be robust and verifiable \citep{raghunathan2018certified,wong2018provable,dvijotham2018training,IBP}.
Our work \svcomment{is the first to} extend incomplete verification \svcomment{to text classification}, considering input perturbations on a simplex and minimising worst case bounds to adversarial attacks in \pswrite{text classification}.
\svcomment{We highlight that the verification of neural networks is an extremely challenging task, and that scaling complete and incomplete methods to large models remains an open challenge.}%

\paragraph{Representations of Combinatorial Spaces.} Word lattices and hypergraphs are data structures that have often been used to efficiently represent and process exponentially large numbers of sentences without exhaustively enumerating them. Applications include automatic speech recognition (ASR) output rescoring~\citep{liu:2016}, machine translation of ASR outputs~\citep{Bertoldi2007SpeechTB}, paraphrase variants~\citep{onishi:2010}, and word segmentation alternatives~\citep{dyer:2008}. The specifications used to characterise the space of adversarial attacks are likewise a compact representation, and the algorithms discussed below operate on them without exhaustive enumeration.

\section{Methodology}
\label{sec:methods}

We assume a fixed initial vector representation $\z_0$ of a given input sentence $z$\jwwrite{\footnote{For brevity, we will refer both to the original symbol sequence and its corresponding vector representation with the same variable name, distinguishing them by styling.}} (e.g.~the concatenation of pretrained word embeddings) 
and use 
a neural network model, i.e.~a series of differentiable transformations $h_k$:
\begin{equation}
\z_k = h_k(\z_{k - 1}) \quad k = 1, \ldots, K
\end{equation}
where $\z_k$ is the vector of activations in the $k$-th layer and the final output $\z_K$ consists of the logits for each class. %
Typically each $h_k$ will be an affine transformation followed by an activation function (e.g. ReLU or sigmoid). The affine transformation can be a convolution (with the inputs and outputs having an implied 2D structure) of a vector of activations at each point in a sequence; in what follows these activations will be concatenated along the sequence to form a vector $\z_k$.

\subsection{Verification}
Verification is the process of examining whether the output of a model satisfies a given specification. %
Formally, this means establishing whether the following holds true for a given \emph{\pswrite{normal}} model input $\x_0$:
 $\forall \z_0 \in \mathcal{X}_\mathrm{in}(\x_0):~ \z_K \in \mathcal{X}_\mathrm{out}$,
where $\mathcal{X}_\mathrm{out}$ characterizes a constraint on the outputs, and $\mathcal{X}_\mathrm{in}(\x_0)$ defines a neighbourhood of $\x_0$ throughout which the constraint should be satisfied.

In our concrete use case, we consider a specification of robustness against adversarial attacks which are defined by bounded input perturbations (synonym flips up to $\delta$ words, or character flips up to $\delta$ characters) of the original sentence $x$. The attack space $\mathcal{X}_\mathrm{in} (\x_0)$ is the set of vector representations (embeddings) of all such perturbed sentences. 
Denoting by $z_{K,y}$ the logit of label $y$, we formulate the output constraint that for all classes $y: z_{K,\y} \geq z_{K,y}$.
This specification establishes that the prediction of \emph{all} perturbed sentences $\z_0 \in \mathcal{X}_\mathrm{in}(\x_0)$ should correspond to the correct label \y.
This specification may equivalently be formulated as a set of half-space constraints on the logits: for each class $y$
\begin{equation}%
    (\onehot{y} - \onehot{\y})\T \z_K \leq 0 \quad \forall \z_0 \in \mathcal{X}_\mathrm{in}(\x_0)
\label{eq:adversarial_spec}%
\end{equation}
where \onehot{i} is a one-hot vector with 1 in the $i$-th position. 
In other words, the true class logit should be greater or equal than those for all other classes $y$, which means the prediction remains constant.
\subsection{Verification as Optimisation}%
Verifying the specification in Eq.~(\ref{eq:adversarial_spec}) can be done by solving the following constrained optimisation problem to find the input that would most strongly violate it: %
\begin{equation}
\begin{aligned}%
\underset{\z_0 \in \mathcal{X}_\mathrm{in}(\x_0)}{\maximize} & \quad \cvec\T \z_K  \\
\text{subject to} & \quad \z_k = h_k(\z_{k - 1}) \quad k = 1, \ldots, K
\end{aligned}%
\label{eq:verification}
\end{equation}
where $\cvec$ is a vector with entries $c_y = 1$, $c_{\y} = -1$ and 0 everywhere else.
If the optimal value of the above optimisation 
problem is smaller than 0, then the specification in Eq.~(\ref{eq:adversarial_spec}) is satisfied, otherwise a counter-example has been found. In our case, this corresponds to a successful adversarial attack.

\subsection{\svcomment{Modeling Input Perturbations using Simplices}}%
\label{sec:simplex}
In the interests of computational feasibility, we will actually attempt to verify the specification on a larger, but more tractable input perturbation space $\bar{\mathcal{X}}_\mathrm{in} \supseteq \mathcal{X}_\mathrm{in}$. Any data point that is verifiable on this larger input perturbation space is necessarily verifiable with respect to the original specification.

In the domain of image classification, $\mathcal{X}_\mathrm{in}$ is often modeled as an $L_\infty$-ball, corresponding to input perturbations in which each pixel may be independently varied within a small interval. However, using such interval bounds is unsuitable for our situation of perturbations consisting of a small number $\delta$ of symbol substitutions.
Although we could construct an axis-aligned bounding box $\bar{\mathcal{X}}_\mathrm{in}$ in embedding space that encompasses all of $\mathcal{X}_\mathrm{in}$, it would over-approximate the perturbation space to such an extent that it would contain perturbations where \emph{all} symbols in the sentence have been substituted simultaneously. %

To remedy this, we propose a tighter over-approximation in the form of a `simplex' in embedding space. 
We first define this for the special case $\delta=1$, in which $\mathcal{X}_\mathrm{in} = \{\x_0\}\cup\{\p^{(m)}_0 : 1\le m\le M\}$ consists of the representations of all $M$ sentences $p^{(m)}$ derived from $x$ by performing a \emph{single} synonym (or character) substitution, together with the unperturbed sentence $x$ itself. 
In this case we define $\bar{\mathcal{X}}_\mathrm{in}$ to be the convex hull $\mathcal{S}_1$ of $\mathcal{X}_\mathrm{in}$.
\pswrite{Note we are not considering contextual embeddings \cite{peters2018deep} here. Each `vertex' $\p^{(m)}_0$ is a sequence of embedding vectors that differs from $\x_0$ at only one word (or character) position.}

For a larger perturbation radius $\delta>1$, the cardinality of $\mathcal{X}_\mathrm{in}$ grows exponentially, so manipulating its convex hull becomes infeasible. However, dilating $\mathcal{S}_1$ centered at $\x_0$, scaling it up by a factor of $\delta$, yields a simplex $\mathcal{S}_\delta$ with $M+1$ vertices that contains $\mathcal{X}_\mathrm{in}$.

More formally, we define a region in the input embedding space based on the $M$ `elementary' perturbations %
$\{\p^{(m)}_0: m = 1 \ldots M\}$ 
of $\x_0$ defined earlier for the $\delta=1$ case.
For perturbations of up to $\delta$ substitutions, we define $\bar{\mathcal{X}}_\mathrm{in}(\x_0)$ as the convex hull of $\{\z^{(m)}_0: m = 0 \ldots M\}$, where $\z^{(0)}_0=\x_0$ denotes the original (unperturbed) sentence representation and, for $m\ge 1$, $\z^{(m)}_0 = \x_0+\delta \cdot (\p^{(m)}_0-\x_0)$.
The convex hull is an over-approximation of $\mathcal{X}_\mathrm{in}(\x_0)$: 
it contains the representations of all sentences derived from $x$ by performing up to $\delta$ substitutions at distinct word (or character) positions.

\subsection{\svcomment{Interval Bound Propagation}}
\label{sec:ibp}
\pswrite{To estimate the optimal value of the problem \eqref{eq:verification}, given an input $\z_0$, we can propagate the upper/lower bounds on the activations $\z_k$ of each layer using interval arithmetic \cite{IBP}.  
}

We begin by computing interval bounds on the first layer's activations. Recall that any input $\z_0 \in \mathcal{X}_\mathrm{in}$ will lie within the convex hull of certain vertices $\{\z^{(m)}_0: m = 0 \ldots M\}$. Then, assuming that the first layer $h_1$ is an affine transformation (e.g. linear or convolutional) followed by a monotonic activation function, the lower and upper bounds on the components $z_{1,i}$ of the first layer's activations $\z_1$ are as follows:
\begin{equation}
\begin{aligned}
\lowerbound{z}{1,i} = \!\!\underset{m = 0, \ldots, M}{\min}\!\! \onehot{i}\T h_1(\z^{(m)}_0) \\
\upperbound{z}{1,i} = \!\!\underset{m = 0, \ldots, M}{\max}\!\! \onehot{i}\T h_1(\z^{(m)}_0)
\end{aligned}
\label{eq:bound_propagation_1}
\end{equation}
Note that these bounds are efficient to compute (by passing each perturbation $\z^{(m)}_0$ through the first layer); in particular there is no need to compute the convex hull polytope.

For subsequent layers $k>1$, the bounds on the components $z_{k,i}$ of $\z_k$ are:
\begin{equation}
\begin{aligned}
\lowerbound{z}{k,i} = \!\!\!\!\!\!\!\!\underset{\lowerbound{\z}{k-1} \leq \z_{k-1} \leq \upperbound{\z}{k-1}}{\min}\!\! \onehot{i}\T h_k(\z_{k - 1}) \\
\upperbound{z}{k,i} = \!\!\!\!\!\!\!\!\underset{\lowerbound{\z}{k-1} \leq \z_{k-1} \leq \upperbound{\z}{k-1}}{\max}\!\! \onehot{i}\T h_k(\z_{k - 1})
\end{aligned}
\label{eq:bound_propagation_2}
\end{equation}

The above optimisation problems can be solved in closed form quickly for  affine layers and monotonic activation functions, as illustrated in \newcite{IBP}. 
Finally, the lower and upper bounds of the output logits $\z_K$ can be used to construct an upper bound on the solution of \eqref{eq:verification}:
\begin{equation}
\underset{\lowerbound{\z}{K} \leq \z_K \leq \upperbound{\z}{K}}{\maximize} \cvec\T \z_K
\label{eq:upper_bound_verification}
\end{equation}

\paragraph{Verifiable Training.}%
The upper bound in \eqref{eq:upper_bound_verification} is fast to compute (only requires two forward passes for upper and lower bounds through the network). Hence, we can define a loss to optimise models such that the models are trained to be verifiable.
Solving~\eqref{eq:upper_bound_verification} is equivalent to finding the worst-case logit difference, and this is achieved when the logit of the true class is equal to its lower bound, and all other logits equal to their upper bounds. Concretely, for each class $y \neq \y$:  $\hat{\z}_{K,y}(\delta) = \upperbound{\z}{K,y}$, and $\hat{\z}_{K,\y}(\delta) = \lowerbound{\z}{K,\y}$.
The training loss can then be formulated as
\begin{equation}
    L = \kappa \underbrace{\ell(\z_K, \y)}_{L_\textrm{\pswrite{normal}}} + (1 - \kappa) \underbrace{\ell(\hat{\z}_K(\delta), \y)}_{L_\textrm{spec}}
\label{eq:loss}
\end{equation}
where $\ell$ is the cross-entropy loss, $\kappa$ a hyperparameter that controls the relative weights between the classification loss $L_\textrm{\pswrite{normal}}$ and specification loss $L_\textrm{spec}$. %
If $\delta = 0$ then $\z_K = \hat{\z}_K(\delta)$, and thus $L$ reduces to a standard classification loss.
Empirically, we found that a curriculum-based training, starting with $\kappa$=1 and linearly decreasing to 0.25, is effective for verifiable training.

\section{Experiments}%
\label{sec:experiments}
\pswrite{We conduct verification experiments on two text classification datasets, Stanford Sentiment Treebank (SST) \cite{socher2013recursive} and AG News corpus, processed in \cite{zhang2015character}.
}
\pswrite{We focus on word-level and character-level experiments on SST and character-level experiments on AG News.}
Our specification is that models should preserve their prediction against up to $\delta$ synonym substitutions or character typos, respectively.

\pswrite{
\subsection{A Motivating Example}}
We provide an example from Table \ref{table:adv_examples} to highlight different evaluation metrics and training methods. 
Given a sentence, ``you ' ve seen them a million times .'', that is predicted correctly (called {\it Nominal Accuracy\footnote{We use the term ``nominal accuracy'' to indicate the accuracy under various adversarial perturbations is much lower.
}}) 
by a classification model, we want to further examine whether the model is robust against character typos (e.g., up to $\delta=3$ typos) to this example.
One way is to use some heuristic to search for a valid example with up to 3 typos that can change the prediction the most (called {\it adversarial example}). 
We evaluate the model using this adversarial example and report the performance (called {\it Adversarial Accuracy}).
However, even if the adversarial example is predicted correctly, one can still ask: is the model truly robust against \emph{any} typos (up to 3) to this example?
In order to have a certificate that the prediction will not change under any $\delta=3$ character typos (called {\it verifiably robust}), we could in theory exhaustively search over all possible cases and check whether any of the predictions is changed (called {\it Oracle Accuracy}).
\jwwrite{If we only allow a character to be replaced by another character nearby on the keyboard, already for this short sentence we need to exhaustively search over 2,951 possible perturbations.}
To avoid this combinatorial growth, we can instead model all possible perturbations using the proposed simplex bounds and propagate the bounds through IBP at the cost of two forward passes. 
Following Eq. \eqref{eq:verification}, we can check whether this example can be verified to be robust against all perturbations (called {\it IBP-Verified Accuracy}). 

There are also a number of ways in which the training procedure can be enhanced to improve the verified robustness of a model against typos to the sentence. The baseline is to train the model with the original/normal sentence directly (called {\it Normal Training}).
Another way is to randomly sample typo sentences among the 2,951 possible perturbations and add these sentences to the training data (called {\it Data Augmentation Training}). 
Yet another way is to find, at each training iteration, the adversarial example among the (subset of) 2,951 possible perturbations that can change the prediction the most; we then use the adversarial example alongside the training example (called {\it Adversarial Training}).
Finally, as simplex bounds with IBP is efficient to run, we can train a model to be verifiable by minimising Eq. \eqref{eq:loss} (called {\it Verifiable Training}).

\subsection{Baselines}
In this section we detail our baseline models.

\paragraph{Adversarial Training.}
In adversarial training \citep{madry2017towards, goodfellow2014explaining}, 
the goal is to optimise the saddle point problem:
\begin{equation}
\label{eq:minmax}
	\min_{\mathbf{\theta}} \mathop{\mathbb{E}}_{(\x_0,y)}\left[\max_{\z_0 \in \mathcal{X}_\mathrm{in}(\x_0)}
    \ell_{\mathbf{\theta}}(\z_0, y)\right] \;
\end{equation}
where the inner maximisation problem is to find an adversarial perturbation $\z_0\in \mathcal{X}_\mathrm{in}(\x_0)$ that can maximise the loss. In the inner maximisation problem, we use HotFlip \cite{ebrahimi2018hotflip} with perturbation budget $\delta$ to find the adversarial example. 
The outer minimisation problem aims to update model parameters such that the adversarial risk of \eqref{eq:minmax} is minimised.
To balance between the adversarial robustness and nominal accuracy, we use an interpolation weight of 0.5 between the original cross-entropy loss and the adversarial risk.

\paragraph{Data Augmentation Training.}
In the data augmentation setup, we randomly sample a valid perturbation $z$ with perturbation budget $\delta$ from a \pswrite{normal} input $x$, and minimise the cross-entropy loss given the perturbed sample $z$ (denoted as data augmentation loss). We also set the interpolation weight between the data augmentation loss and the original \pswrite{normal} cross-entropy loss to 0.5.

\paragraph{Normal Training.} 
In normal training, we use the likelihood-based training using the normal training input $x$.

\begin{table*}[h]
\centering
\resizebox{\textwidth}{!}{%
\begin{tabular}{l|ccc|ccc|ccc}
\toprule
&&{\bf SST-Char-Level}&&&{\bf SST-Word-Level}&&&{\bf AG-Char-Level}\\\hline
{\bf Training} & {\bf Acc.} & {\bf Adv. Acc.} & {\bf Oracle} &  {\bf Acc.} & {\bf Adv. Acc.} & {\bf Oracle} &  {\bf Acc.} & {\bf Adv. Acc.} & {\bf Oracle}
\\\hline

Normal    & 79.8 & 36.5 & 10.3 & 84.8 & 71.3 & 69.8 & 89.5 & 75.4 & 65.1 \\
Adversarial &  79.0 & {\bf 74.9} & 25.8 & {\bf 85.0} & 76.8 & 74.6 & {\bf 90.5} & 85.5 & 81.6 \\
Data aug.   & 79.8 & 37.8 & 13.7 & 85.4 & 72.7 & 71.6 & 88.4 & 77.5 & 72.0 \\
Verifiable (IBP)  &  74.2 & 73.1 & {\bf 73.1} & 81.7 & {\bf 77.2} & {\bf 76.5} & 87.6 & {\bf 87.1} & {\bf 87.1} \\
\bottomrule
\end{tabular}
}\vspace{-.5mm}
\caption{{Experimental results for changes up to $\delta$=3 and $\delta$=2 symbols on SST and AG dataset, respectively. We compare normal training, adversarial training, data augmentation and IBP-verifiable training, using three metrics on the test set: the nominal accuracy, adversarial accuracy, and exhaustively verified accuracy (Oracle) (\%).}}
\label{tab:word-level}
\end{table*}
\begin{table*}[ht]
\centering
\scalebox {0.87} {

\begin{tabular}{c|l}
\hline
Prediction & SST word-level examples (by exhaustive verification, not by adversarial attack)  \\\hline

+ & it ' s the kind of pigeonhole-resisting romp that hollywood too rarely provides .\\
- & it ' s the kind of pigeonhole-resisting romp that hollywood too rarely {\bf gives} .\\\hline
- & sets up a nice concept for its fiftysomething leading ladies , but fails loudly in execution .\\
+ & sets up a nice concept for its fiftysomething leading ladies , but fails {\bf aloud} in execution .\\
\hline
Prediction & SST character level examples (by exhaustive verification, not by adversarial attack) \\\hline
- & you ' ve seen them a million times . \\
+ & you ' ve se{\bf r}n them a million times .\\\hline
+ & choose your reaction : a. ) that sure is funny ! \\
- & choose {\bf t}our reaction : a. ) that sure is funny !\\ 
\hline
\end{tabular}
}\vspace{-.5mm}
\caption{Pairs of original inputs and adversarial examples for SST sentiment classification found via an exhaustive verification oracle, but not found by the HotFlip attack (i.e., the HotFlip attack does not change the model prediction). The bold words/characters represent the flips found by the adversary that change the predictions.}
\label{table:adv_examples}
\end{table*}

\subsection{Setup}
\pswrite{
We use a shallow convolutional network with a small number of fully-connected layers for SST and AG News experiments.
The detailed model architectures and hyperparameter details are introduced in the supplementary material. 
Although we use shallow models for ease of verifiable training, our nominal accuracy is on par with previous work such as \citet{socher2013recursive}~(85.4\%) and \citet{P18-1041}~(84.3\%) in SST and \citet{zhang2015character}~(87.18\%) in AG News.
During training, we set the maximum number of perturbations to $\delta=3$, and evaluate performance with the maximum number of perturbations from $\delta=1$ to $6$ at test time.
}

For word-level experiments, we construct the synonym pairs using the PPDB database \cite{ganitkevitch2013ppdb} and filter the synonyms with fine-grained part-of-speech tags using Spacy \cite{spacy2}. 
For character-level experiments, we use synthetic keyboard typos from \citet{Belinkov_Bisk2017}, and allow one possible alteration per character that is adjacent to it on an American keyboard. %
The allowable input perturbation space is much larger than for word-level synonym substitutions, as shown in Table \ref{tab:worst_case_perturbations}.

\subsection{Evaluation Metrics}
We use the following four metrics to evaluate our models:
i) test set accuracy (called Acc.), 
ii) adversarial test accuracy (called Adv. Acc.), which uses samples generated by HotFlip attacks on the original test examples,
iii) verified accuracy under IBP verification (called IBP-verified), that is, the ratio of test samples for which IBP can verify that the specification is not violated, and
iv) exhaustively verified accuracy (called Oracle), computed by enumerating all possible perturbations given the perturbation budget $\delta$, where a sample is verifiably robust if the prediction is unchanged under all valid perturbations.

\begin{figure*}[ht]
\centering
\begin{subfigure}{.26\textwidth}
\includegraphics[width=1.04\linewidth]{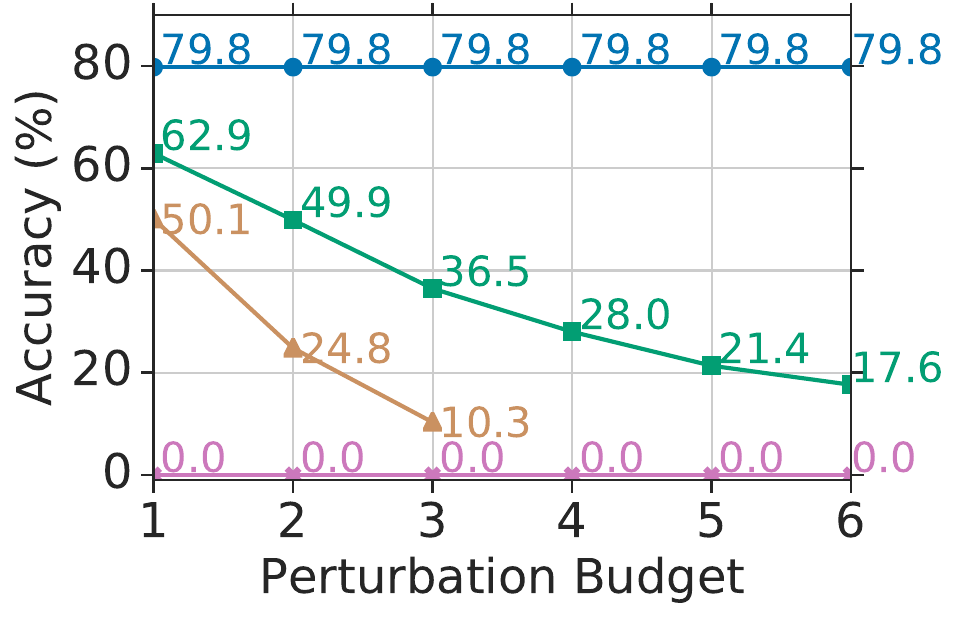}
 \caption{\pswrite{Normal} Training}
  \label{fig:cfig1}
\end{subfigure}%
\begin{subfigure}{.43\textwidth}
\hspace*{10mm}
\includegraphics[width=1.04\linewidth]{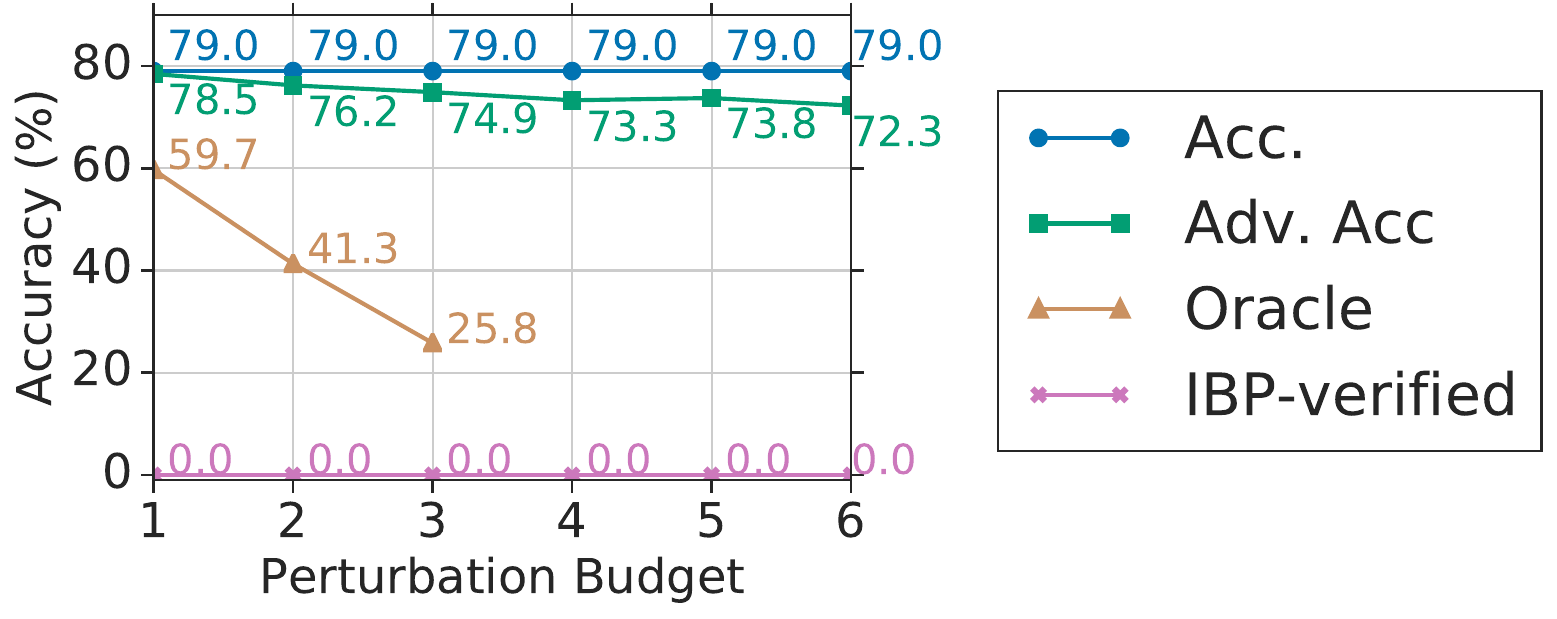}
  \caption{Adversarial Training}
  \label{fig:cfig2}
  \end{subfigure}%
  \hfill
  \begin{subfigure}{.26\textwidth}
\includegraphics[width=1.04\linewidth]{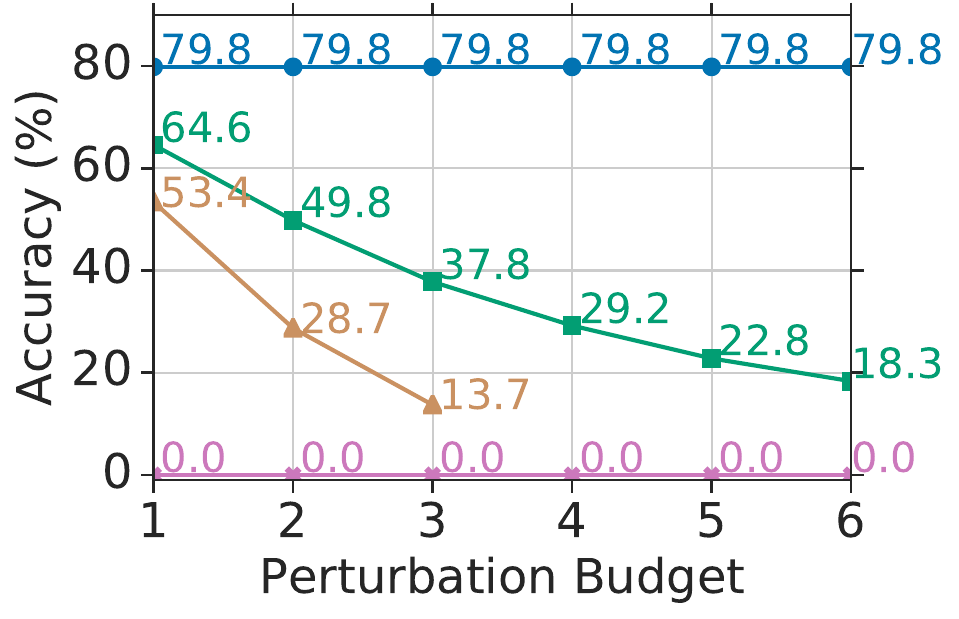}
  \caption{Data Augmentation Training}
  \label{fig:cfig3}
  \end{subfigure}%
    \begin{subfigure}{.43\textwidth}
\hspace*{10mm}
\includegraphics[width=1.04\linewidth]{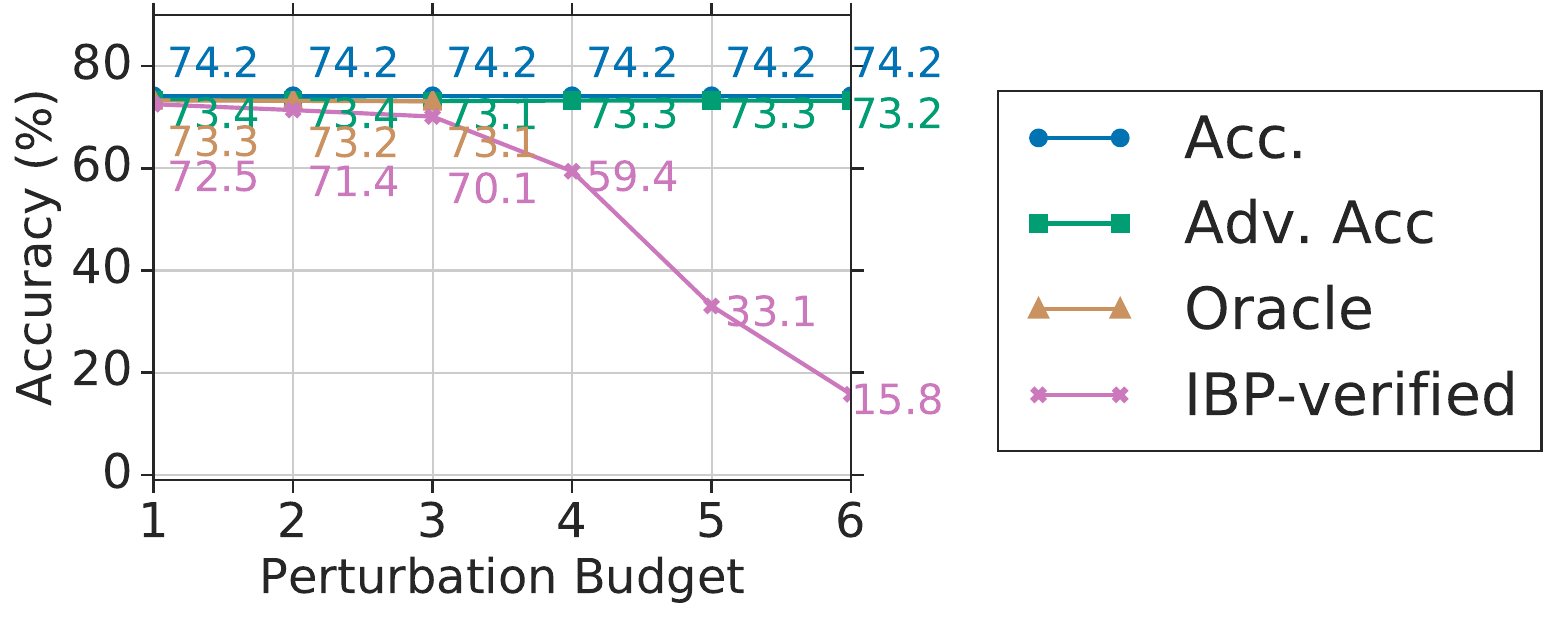}%

  \caption{Verifiable Training (IBP)}
  \label{fig:cfig4}
  \end{subfigure}%
\caption{{
SST character-level models with different training objectives (trained at $\delta$=$3$) against different perturbation budgets in nominal accuracy, adversarial accuracy, exhaustively verified accuracy (Oracle), and IBP verified accuracy. Note that exhaustive verification is not scalable to perturbation budget 4 \jwwrite{and beyond}.}}
\label{fig:char_level_detailed_results}
\end{figure*}

\begin{figure*}[h!]
\centering
\begin{subfigure}{.26\textwidth}
  \centering
\includegraphics[width=1.04\linewidth]{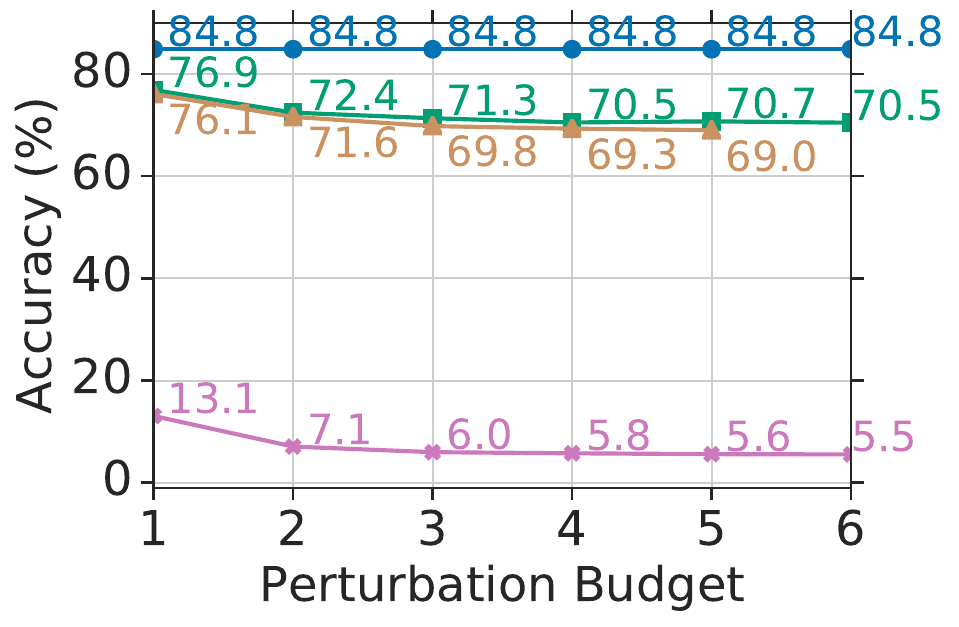}
  \caption{\pswrite{Normal} Training \emph{(GloVe)}}
  \label{fig:wfig1}
\end{subfigure}%
    \begin{subfigure}{.43\textwidth}
\hspace*{10mm}
\includegraphics[width=1.04\linewidth]{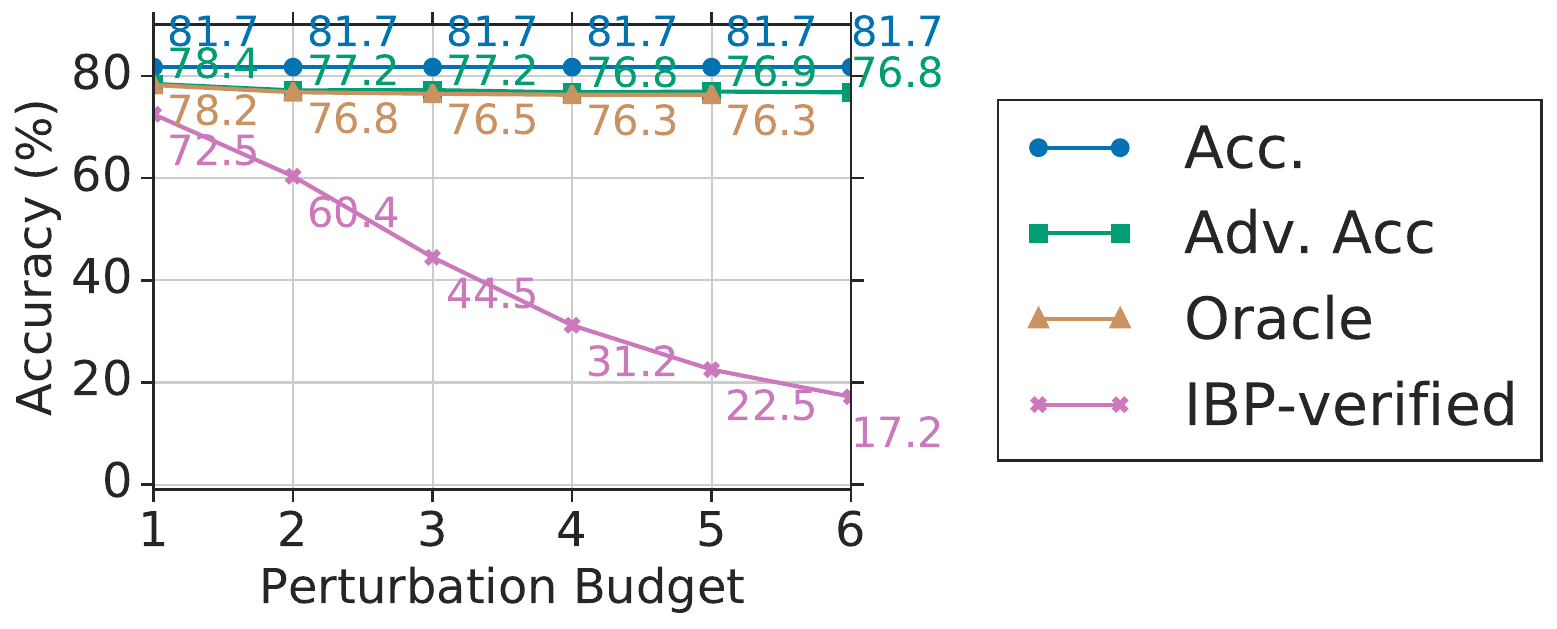}
  \caption{Verifiable Training (IBP) \emph{(GloVe)}}
 \label{fig:wfig4}
\end{subfigure}%

\begin{subfigure}{.26\textwidth}
  \centering
\includegraphics[width=1.04\linewidth]{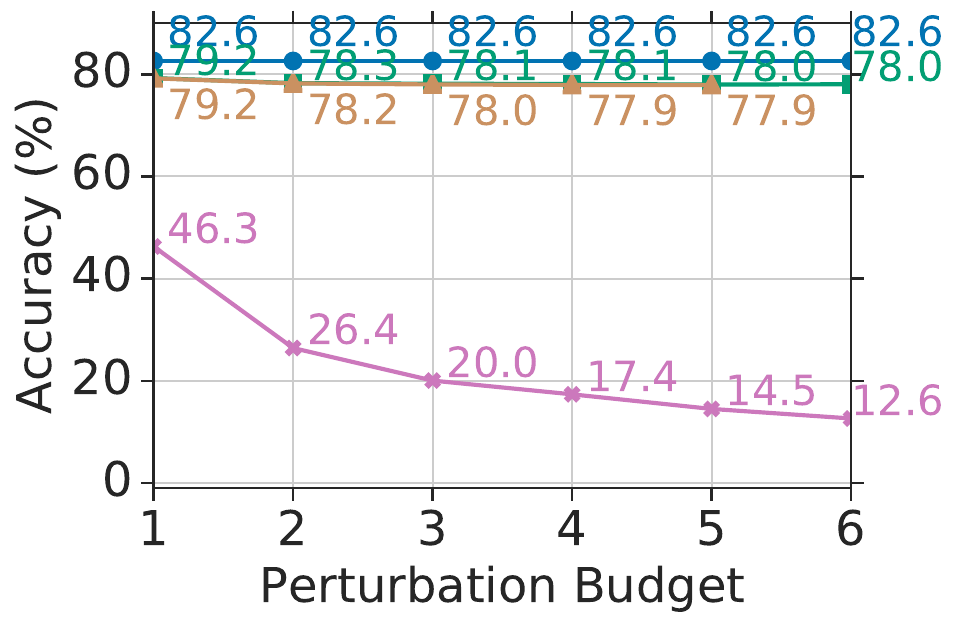}
  \caption{\pswrite{Normal} Training \emph{(CF)}}
  \label{fig:wfig1_counter}
\end{subfigure}%
 \begin{subfigure}{.43\textwidth}
\hspace*{10mm}
\includegraphics[width=1.04\linewidth]{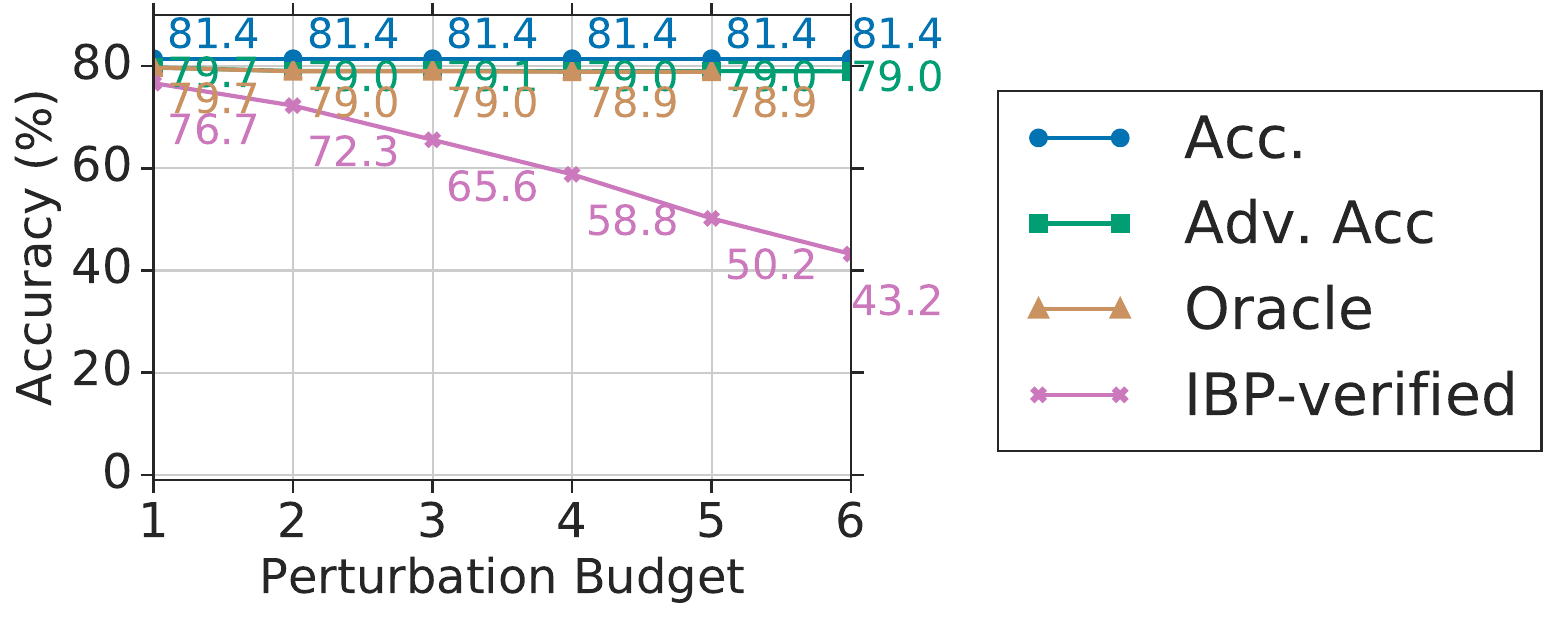}
  \caption{Verifiable Training (IBP) \emph{(CF)}}
 \label{fig:wfig4_counter}
  \end{subfigure}%
\caption{{
SST word-level models with normal and verifiable training objectives (trained at $\delta$=$3$) using \emph{GloVe} and \emph{counter-fitted (CF)} embeddings against different perturbation budgets in nominal accuracy, adversarial accuracy, exhaustively verified accuracy (Oracle), and IBP verified accuracy. Note that exhaustive verification is not scalable to perturbation budget 6\jwwrite{ and beyond}. %
}}
\label{fig:word_level_detailed_results}
\end{figure*}

\subsection{Results}

Table \ref{tab:word-level} shows the results of IBP training and baseline models under $\delta=3$ and $\delta=2$\footnote{Note that the exhaustive oracle is not computationally feasible beyond $\delta=2$ on AG News.} perturbations \pswrite{on SST and AG News, respectively.} %
Figures \ref{fig:char_level_detailed_results} and \ref{fig:word_level_detailed_results} show the character- and word-level results with $\delta$ between 1 and 6 under four metrics on the SST test set; similar figures for SST word-level (adversarial training, data augmentation) models and AG News dataset can be found in the supplementary material.

\paragraph{Oracle Accuracy and Adversarial Accuracy.}
In Table \ref{tab:word-level}, comparing adversarial accuracy with exhaustive verification accuracy (oracle), we observe that although adversarial training is effective at defending against HotFlip attacks (74.9 / 76.8 / 85.5\%), the oracle adversarial accuracy under exhaustive testing (25.8 / 74.6 / 81.6\%) is much lower in SST-character / SST-word / AG-character level, respectively.
For illustration, we show some concrete adversarial examples from the HotFlip attack in Table \ref{table:adv_examples}. For some samples, even though the model is robust with respect to HotFlip attacks, its predictions are incorrect for stronger adversarial examples obtained using the exhaustive verification oracle. 
This underscores the need for verification, as robustness with respect to suboptimal adversarial attacks alone might give a false sense of security.

\paragraph{Effectiveness of \pswrite{Simplex Bounds with IBP}.}
Rather than sampling individual points from the perturbation space, IBP training covers the full space at once.
The resulting models achieve the highest exhaustively verified accuracy at the cost of \jwwrite{only%
}
moderate deterioration in nominal accuracy (Table \ref{tab:word-level}).
At test time, IBP allows for constant-time verification with arbitrary $\delta$, whereas exhaustive verification requires evaluation over an exponentially growing search space.
\paragraph{Perturbation Space Size.}
In Table \ref{tab:word-level}, when the perturbation space is larger (SST character-level vs. SST word-level), 
(a) across models, there is a larger gap in adversarial accuracy and true robustness (oracle);
(b) the difference in oracle robustness between IBP and adversarial training is even larger (73.1\% vs.~25.8\% and 76.5\% vs.~74.6\%). %

\paragraph{Perturbation Budget.}
\label{sec:word-level_perturbations}
In Figures \ref{fig:char_level_detailed_results} and \ref{fig:word_level_detailed_results}, we compare normal training, adversarial training, data augmentation, and verifiable training models with four metrics under various perturbation budgets \pswrite{on the SST dataset}. 
Overall, as the perturbation budget increases, the adversarial accuracy, oracle accuracy, and IBP-verified accuracy decrease.
We can observe that even for large perturbation budgets, verifiably trained models are still able to verify a sizable number of samples. 
Again, although adversarial accuracy flattens for larger perturbation budgets in the word level experiments, oracle verification can further find counterexamples to change the prediction. %
Note that exhaustive verification becomes intractable with large perturbation sizes.%

\ignore{
\paragraph{Ease of Verification (Computation of True Robustness).}
For every training method, we can compute the true robustness using exhaustive verification. 
However, this oracle is extremely computationally expensive (especially in character level perturbations). 
On the other hand, verification via IBP provides a lower bound on the worst-case results, but this is generally loose for arbitrary networks.
IBP-verifiable training succeeds in tightening these bounds and results in much improved rates of IBP-verification at test time, compared to all other training methods.
We furthermore can observe that models trained to become verifiable (with IBP training objective)  achieve better adversarial accuracy and exhaustively verified accuracy, with a small deterioration in nominal accuracy compared to \pswrite{normal} training for character-level models.
}

\paragraph{Computational Cost of Exhaustive Verification.}
\label{sec:exhaustive_verification_comparison}
The perturbation space in NLP problems is discrete and finite, and a valid option to verify the specification is to exhaustively generate predictions for all $\z_0 \in \mathcal{X}_\mathrm{in} (\x_0)$, and then check if at least one does not match the correct label.
Conversely, such an exhaustive (oracle) approach can also identify the strongest possible attack.
But the size of $\mathcal{X}_\mathrm{in}$ grows exponentially with $\delta$, and exhaustive verification quickly becomes prohibitively expensive. 

\begin{figure}[ht]
\centering
\includegraphics[width=.8\linewidth]{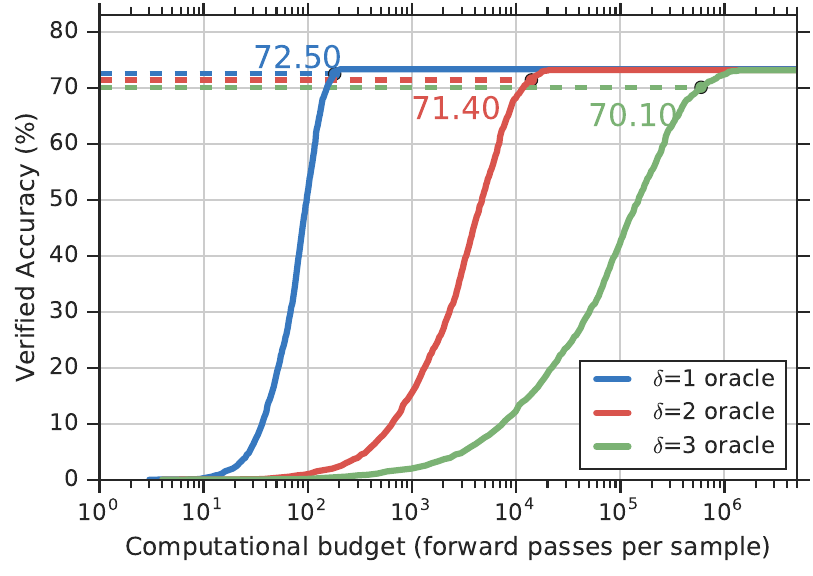}
\caption{{
Verified accuracy vs.~computation budget for exhaustive verification oracles on the SST character-level test set, for an IBP-trained model trained with $\delta$=$3$. Solid lines represent the number of forward passes required to verify a given proportion of the test set using exhaustive search. 
Dashed lines indicate verification levels achievable using IBP verification, which comes at small and constant cost, and is thus orders of magnitude more efficient. 
}}
\label{fig:exhaustive}
\end{figure}

In Table \ref{tab:worst_case_perturbations}, we show the maximum perturbation space size in the SST \pswrite{and AG News} test set for different perturbation radii $\delta$. This number grows exponentially as $\delta$ increases.
To further illustrate this, Figure \ref{fig:exhaustive} shows the number of forward passes required to verify a given proportion of the SST test set for an IBP-trained model using exhaustive verification and IBP verification. 
IBP reaches verification levels comparable to an exhaustive verification oracle,
but requires only two forward passes to verify any sample -- one pass for computing the upper, and one for the lower bounds.~%
Exhaustive verification, on the other hand, requires several orders of magnitude more forward passes, and there is a tail of samples with extremely large attack spaces.

\begin{table}[t]
\centering
\resizebox{.45\textwidth}{!}{%
\begin{tabular}{l r r r}\toprule
Perturbation radius     & $\delta=$1    & $\delta=$2    & $\delta=$3\\\midrule
SST-word              &  49           & 674           & 5,136     \\
SST-character         & 206           & 21,116        & 1,436,026 \\
AG-character      & 722           & 260,282       &  - \\\bottomrule
\end{tabular}}
\caption{Maximum perturbation space size in the SST and AG News test set using word / character substitutions, which is the maximum number of forward passes per sentence to evaluate in the exhaustive verification. %
}
\label{tab:worst_case_perturbations}
\end{table}

\subsection{Counter-Fitted Embeddings}\label{sec:counterfitted}
As shown in Figures \ref{fig:char_level_detailed_results} and \ref{fig:wfig1}, although IBP can verify arbitrary networks in theory, the verification bound is very loose except for models trained to be IBP-verifiable.
One possible reason is the potentially large volume of the perturbation simplex. 
Since representations of substitution words/characters are not necessarily close to those of synonyms/typos in embedding space, the vertices of the simplex could be far apart, and thus cover a large area in representation space.
Therefore, when propagating the interval bounds through the network, the interval bounds become too loose and fail to verify most of the examples if the models are not specifically trained.
To test this hypothesis, we follow \citet{mrksic:2016:naacl} and use fine-tuned GloVe embeddings trained to respect linguistic constraints; these representations (called counter-fitted embeddings) force synonyms to be closer and antonyms to be farther apart using word pairs from the PPDB database \cite{ganitkevitch2013ppdb} and WordNet \cite{WordNet}. 
We repeat the word level experiments with these counter-fitted embeddings, Figures~\ref{fig:wfig1_counter} and \ref{fig:wfig4_counter} show the experimental results. %
We observe that IBP verified accuracy is now substantially higher across models, especially for $\delta=1, 2, 3$. 
The examples which IBP can verify increase by up to 33.2\% when using the counter-fitted embeddings (normal training, $\delta=1$). %
Moreover, adversarial and exhaustively verified accuracy are also improved, at the cost of a mild deterioration in nominal test accuracy.
The IBP-trained model also further improves both its oracle accuracy and IBP verified accuracy. 
These results validate our hypothesis that reducing the simplex volume via soft linguistic constraints can provide even tighter bounds for IBP, resulting in larger proportions of verified samples.

\section{\pswrite{Discussion}}
Our experiments indicate that adversarial attacks are not always the \emph{worst} adversarial inputs, which can only be revealed via verification.
On the other hand, exhaustive verification is computationally very expensive.
Our results show that \pswrite{using the proposed simplex bounds} with IBP can verify a sizable amount of test samples, and can be considered a potent verification method in an NLP context.
 We note however two limitations within the scope of this work: 
i) limited model depth: we only investigated models with few layers. IBP bounds are likely to become looser as the number of layers increases.
ii) limited model types: we only studied models with CNN and fully connected layers.

\jwwrite{
We focused on the HotFlip attack to showcase specification verification in the NLP context, with the goal of understanding factors that impact its effectiveness (e.g.~the perturbation space volume, see Section \ref{sec:counterfitted}). It is worth noting that symbol substitution is general enough to encompass other threat models such as lexical entailment perturbations \cite{glockner-etal-2018-breaking}, and could potentially be extended to the addition of pre/postfixes \cite{jia2017adversarial,trigger:emnlp19}.%
}

Interesting directions of future work include: tightening IBP bounds to allow applicability to deeper models, investigating bound propagation in other types of neural architectures (e.g.~those based on recurrent networks or self-attention), and exploring other forms of specifications in NLP.

\section{Conclusion}

We introduced formal verification of \pswrite{text classification models against synonym and character flip perturbations.} 
Through experiments, 
we demonstrated the effectiveness of \pswrite{the proposed simplex bounds} with IBP both during training and testing, and found  \jwwrite{%
weaknesses} of adversarial training compared with exhaustive verification. 
Verifiably trained models achieve the highest exhaustive verification accuracy on SST and AG News. IBP verifies models in constant time, which is exponentially more efficient than naive verification via exhaustive search.

\bibliography{emnlp2019}
\bibliographystyle{acl_natbib}

\clearpage
\input{appendix.tex}

\end{document}

%% file: appendix.tex
\appendix

\section{Experimental Setup}
\label{sec:experimental_setup}
\subsection{Dataset Statistics}
The SST dataset consists of 67,349 training, 872 validation, and 1,821 test samples with binary sentiment annotations.
The AG News contains 120,000 training and 7,600 test samples with 4 classes.

\subsection{Detailed Setup}%

We select model architectures to achieve a reasonable tradeoff \cite{tsipras2018robustness} between nominal accuracy and robust accuracy using the validation set. 
In the SST word-level experiments, we use a 1-layer convolutional network with 100 kernels of width 5, followed by a ReLU, an average pool, and a linear layer. 
We use pre-trained 300-dimensional GloVe embeddings \cite{pennington2014glove}, and use counter-fitted embeddings \cite{mrksic:2016:naacl} in Section \ref{sec:counterfitted}. The pre-trained word embeddings are fixed during training.
In the SST character-level experiments, we use a 1-layer convolutional network with 100 kernels of width 5, followed by a ReLU, an average pool, followed by a linear layer.
We set the character embedding dimension to 150, randomly initialise them, and fine-tune the embeddings during training. %
\pswrite{In the AG News character-level experiments, we follow the setup in \citet{zhang2015character} using lower-case letters only and truncate the character sequences to have at most 300 characters during training. %
We use a 1-layer convolutional network with 100 kernels of width 10, followed by a ReLU, an average pool, and two fully-connected layers with 100 hidden units, followed by a linear layer.}
We set the character embedding dimension to 150, randomly initialise them, and fine-tune the embeddings during training. %
Note since the proposed technique is efficient, we can scale up to deeper networks for better nominal accuracy at the cost of verified accuracy, as the bounds become looser. 

We use Adam~\citep{kingma2015adam} as our optimisation method, perform early stopping, and tune our hyperparameters (learning rate, loss ratio $\kappa$) on the validation set.

\section{Additional Experimental Results and Discussion}

\subsection{Ease of Verification (Computation of True Robustness)}
For every training method, we can compute the true robustness using exhaustive verification. 
However, this oracle is extremely computationally expensive (especially in character-level perturbations). 
On the other hand, verification via IBP provides a lower bound on the worst-case results, but this is generally loose for arbitrary networks.
IBP-verifiable training succeeds in tightening these bounds and results in much improved rates of IBP-verification at test time, compared to all other training methods.
We furthermore can observe that models trained to become verifiable (with IBP training objective)  achieve better adversarial accuracy and exhaustively verified accuracy, with a small (or no) deterioration in nominal accuracy compared to \pswrite{normal} training. %

\subsection{SST Word Embeddings Comparison}
\begin{figure*}[h!]
\centering
\begin{subfigure}{.28\textwidth}
  \centering
\includegraphics[width=1.02\linewidth]{images/word_nominal_performance.pdf}
  \caption{\pswrite{Normal} Training}
  \label{fig:appendix_wfig1}
\end{subfigure}%
\begin{subfigure}{.45\textwidth}
\hspace*{10mm}
\includegraphics[width=1.02\linewidth]{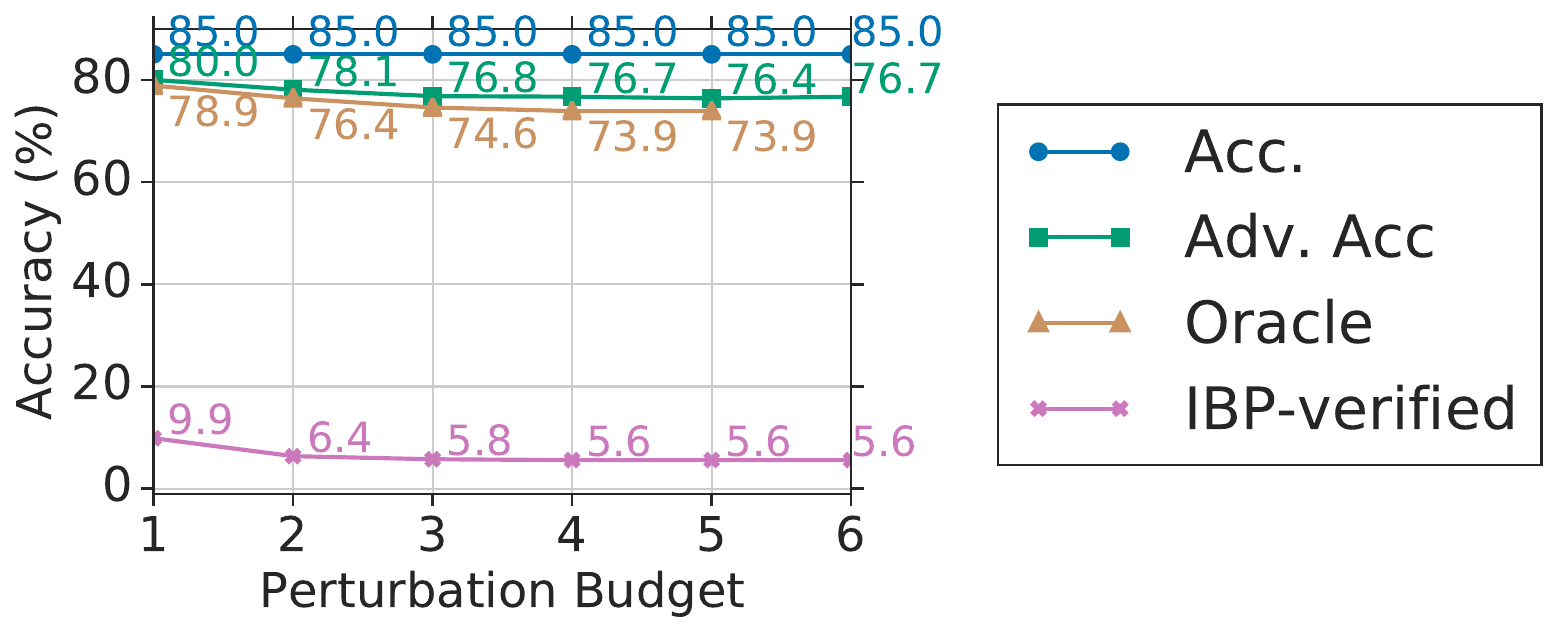}
  \caption{Adversarial Training}
  \label{fig:appendix_wfig2}
  \end{subfigure}%
  \hfill
  \begin{subfigure}{.28\textwidth}
\includegraphics[width=1.02\linewidth]{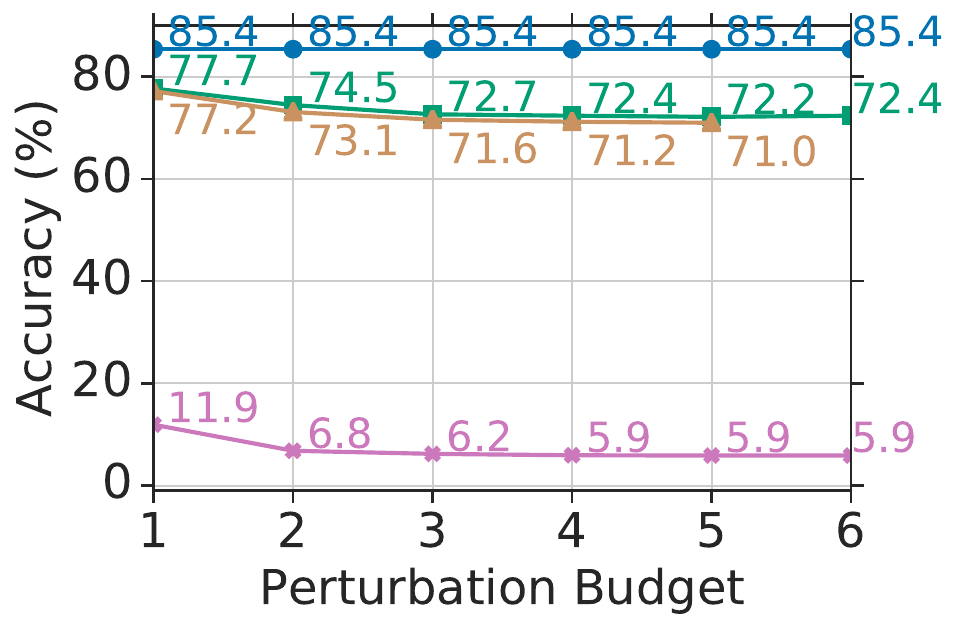}
  \caption{Data Aug. Training}
  \label{fig:appendix_wfig3}
  \end{subfigure}%
    \begin{subfigure}{.45\textwidth}
\hspace*{10mm}
\includegraphics[width=1.02\linewidth]{images/word_verifiable_performance.pdf}
  \caption{Verifiable Training (IBP)}
 \label{fig:appendix_wfig4}
  \end{subfigure}%
\caption{{
SST word-level models with different training objectives (trained at $\delta$=$3$) using \emph{GloVe} embeddings against different perturbation budgets in nominal accuracy, adversarial accuracy, exhaustively verified accuracy (Oracle), and IBP verified accuracy. Note that exhaustive verification is not scalable to perturbation budget 6 and beyond.}}
\label{fig:appendix_word_level_detailed_results}
\end{figure*}

\begin{figure*}[h!]
\centering
\begin{subfigure}{.28\textwidth}
  \centering
\includegraphics[width=1.02\linewidth]{images/counter_fitting_word_nominal_performance.pdf}
  \caption{\pswrite{Normal} Training}
  \label{fig:appendix_wfig1_counter}
\end{subfigure}%
\begin{subfigure}{.45\textwidth}
\hspace*{10mm}
\includegraphics[width=1.02\linewidth]{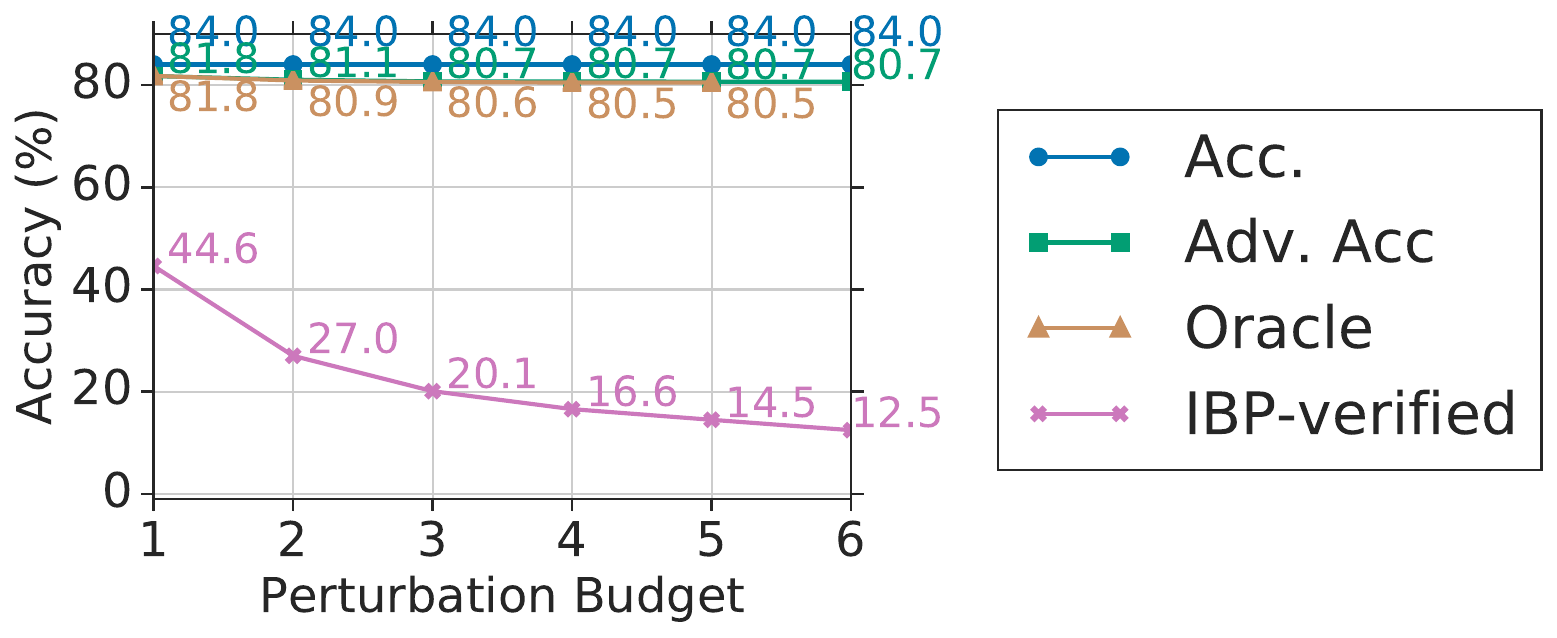}
  \caption{Adversarial Training}
  \label{fig:appendix_wfig2_counter}
  \end{subfigure}%
  \hfill
  \begin{subfigure}{.28\textwidth}
\includegraphics[width=1.02\linewidth]{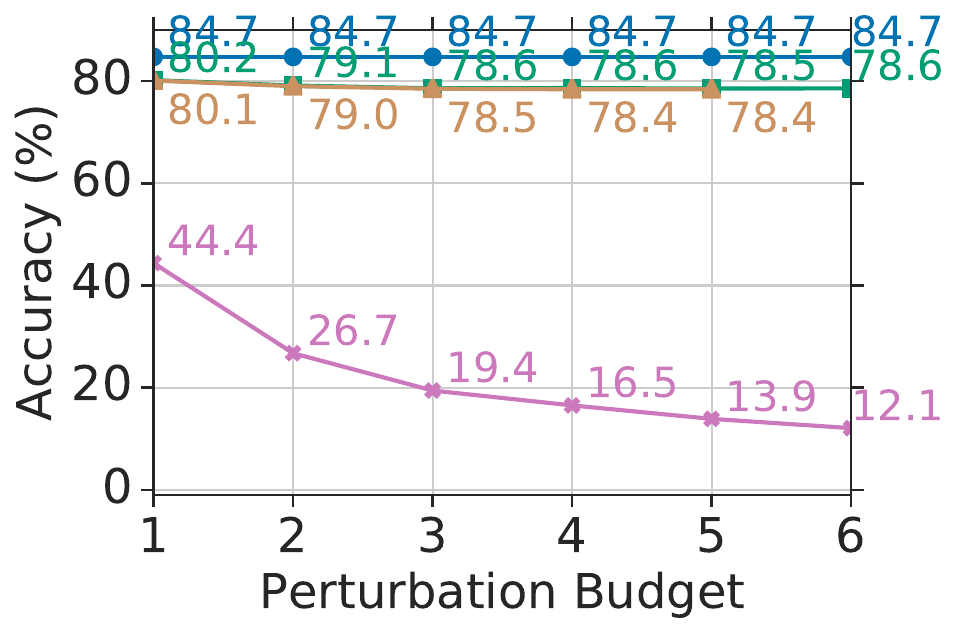}
  \caption{Data Aug. Training}
  \label{fig:appendix_wfig3_counter}
  \end{subfigure}%
    \begin{subfigure}{.45\textwidth}
\hspace*{10mm}
\includegraphics[width=1.02\linewidth]{images/counter_fitting_word_verifiable_performance.pdf}
  \caption{Verifiable Training (IBP)}
 \label{fig:appendix_wfig4_counter}
  \end{subfigure}%
\caption{{
SST word-level models (trained at $\delta$=$3$) using \emph{counter-fitted} embeddings against different perturbation budgets in nominal accuracy, adversarial accuracy, exhaustively verified accuracy (Oracle), and IBP verified accuracy. Note that exhaustive verification is not scalable to perturbation budget 6 and beyond.}}
\label{fig:appendix_counter_fitting_word_level_detailed_results}
\end{figure*}

In Figures \ref{fig:appendix_word_level_detailed_results} and \ref{fig:appendix_counter_fitting_word_level_detailed_results}, we show the experimental results of different models and metrics using GloVe and counter-fitted embeddings, respectively. 


\subsection{AG News}
\begin{figure*}[ht]
\centering
\begin{subfigure}{.28\textwidth}
\includegraphics[width=1.02\linewidth]{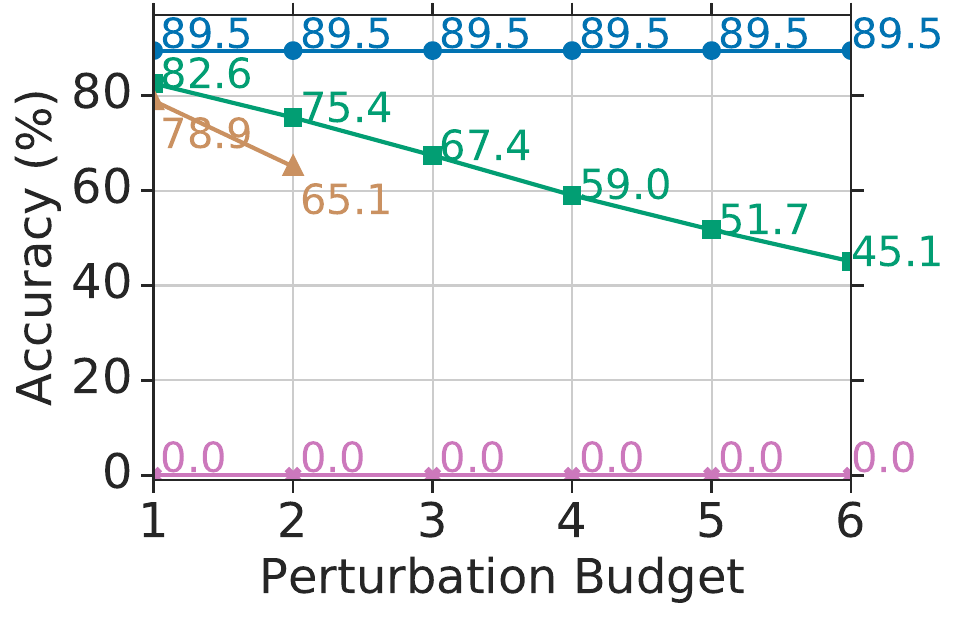}
 \caption{\pswrite{Normal} Training}
  \label{fig:ag_cfig1}
\end{subfigure}%
\begin{subfigure}{.45\textwidth}
\hspace*{10mm}
\includegraphics[width=1.02\linewidth]{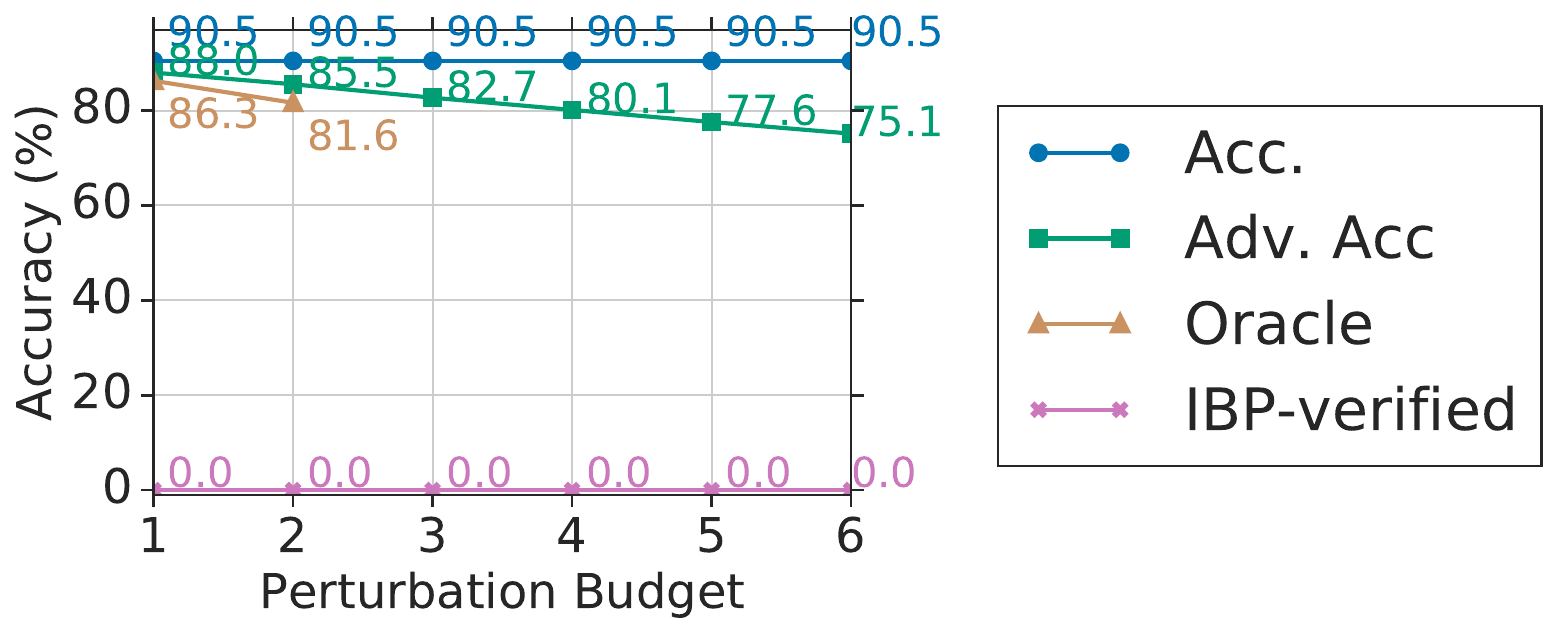}
  \caption{Adversarial Training}
  \label{fig:ag_cfig2}
  \end{subfigure}%
  \hfill
  \begin{subfigure}{.28\textwidth}
\includegraphics[width=1.02\linewidth]{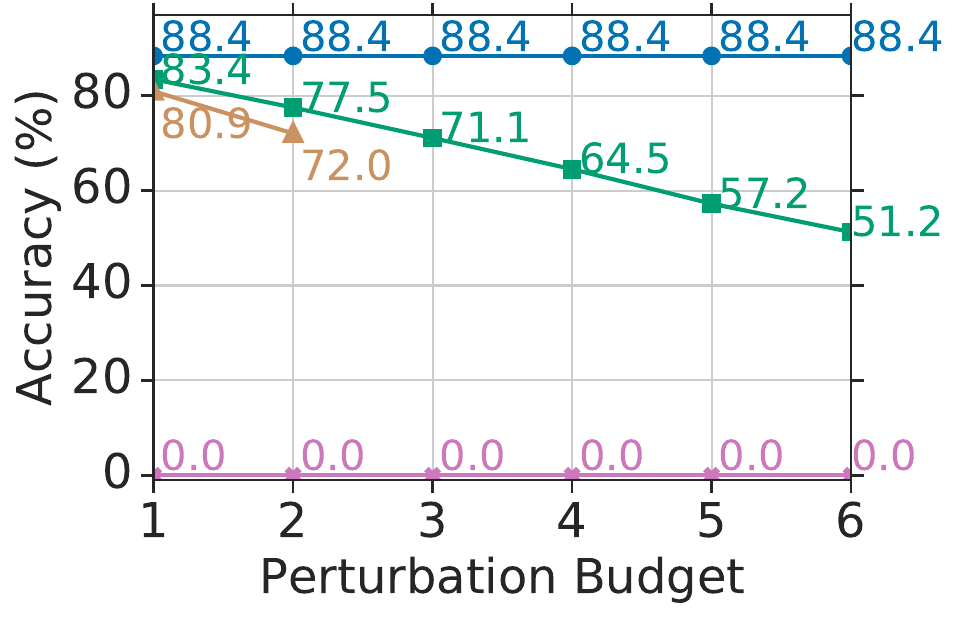}
  \caption{Data Augmentation Training}
  \label{fig:ag_cfig3}
  \end{subfigure}%
    \begin{subfigure}{.45\textwidth}
\hspace*{10mm}
\includegraphics[width=1.02\linewidth]{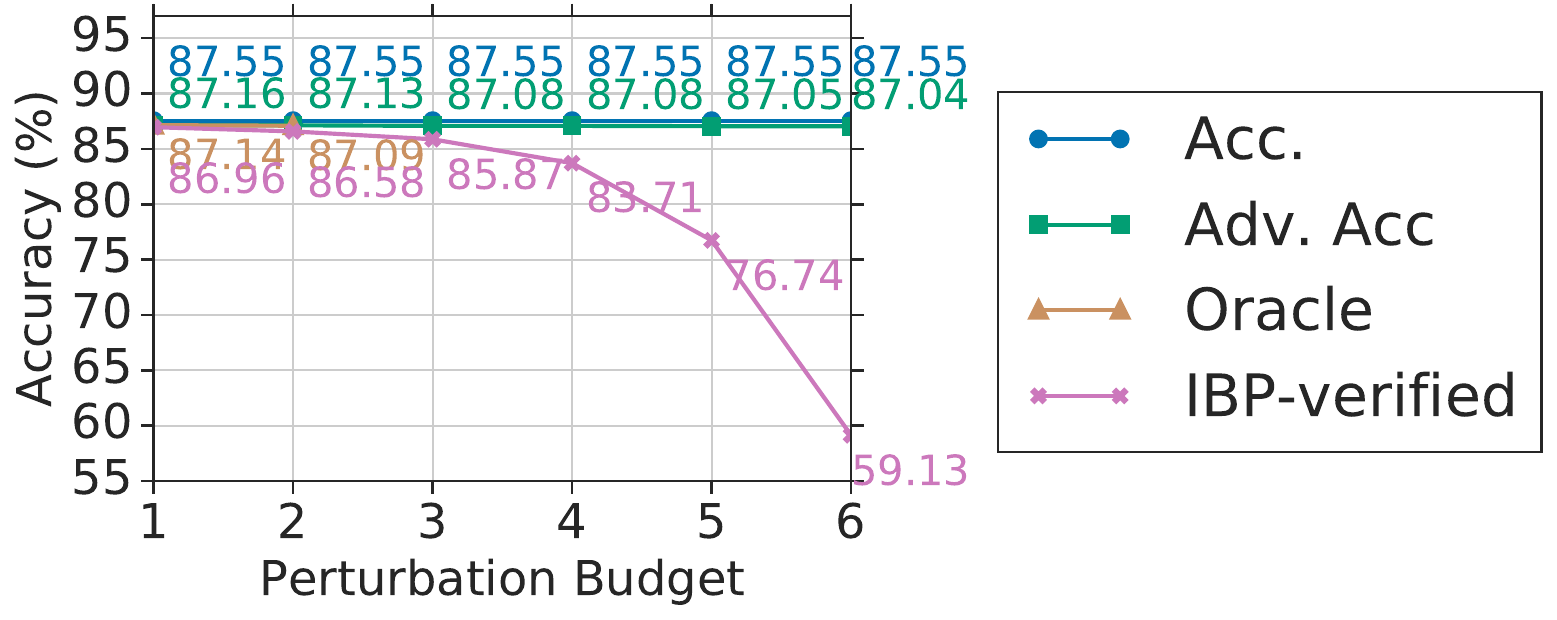}
  \caption{Verifiable Training (IBP)}
  \label{fig:ag_cfig4}
  \end{subfigure}%
\caption{{
AG News character-level models with different training objectives (trained at $\delta$=$3$) against different perturbation budgets in nominal accuracy, adversarial accuracy, exhaustively verified accuracy (Oracle), and IBP verified accuracy. Note that exhaustive verification is not scalable to perturbation budget 3 and beyond.}}
\label{fig:agnews_char_level_detailed_results}
\end{figure*}

\pswrite{In Figure \ref{fig:agnews_char_level_detailed_results}, we compare normal training, adversarial training, data augmentation, and verifiable training models with four metrics under various perturbation budgets on the AG News dataset at the character level.}
In Figure \ref{fig:ag_cfig4}, our verifiable trained model achieves not only the strongest adversarial and oracle accuracy, but achieves very tight bounds with respect to the oracle results. 
Note IBP verification only requires 2 forward passes to verify any examples, whereas oracle evaluation (exhaustive search) uses up to 260,282 forward passes for examining a single example at $\delta=2$. 
